%% file: PaperForReview.tex
\documentclass[10pt,twocolumn,letterpaper]{article}

\usepackage{cvpr}              

\usepackage{graphicx}
\usepackage{amsmath}
\usepackage{amssymb}
\usepackage{booktabs}

\usepackage[table,xcdraw]{xcolor}
\usepackage{subcaption}
\usepackage[symbol]{footmisc}

\usepackage[pagebackref,breaklinks,colorlinks]{hyperref}

\usepackage[capitalize]{cleveref}
\crefname{section}{Sec.}{Secs.}
\Crefname{section}{Section}{Sections}
\Crefname{table}{Table}{Tables}
\crefname{table}{Tab.}{Tabs.}

\iftrue
\newcommand{\xcyan}[1]{\textcolor{cyan}{#1}}
\else
\newcommand{\xcyan}[1]{}
\fi

\input{preamble.tex}
\input{math_commands.tex}

\def\cvprPaperID{6052} 
\def\confName{CVPR}
\def\confYear{2023}

\begin{document}

\title{NeRDi: Single-View NeRF Synthesis \\ with Language-Guided Diffusion as General Image Priors}

\author{Congyue Deng$^2$\footnotemark
\quad Chiyu ``Max'' Jiang$^1$ \quad Charles R. Qi$^1$ \quad Xinchen Yan$^1$ \quad Yin Zhou$^1$ \\
Leonidas Guibas$^{2,3}$ \quad Dragomir Anguelov$^1$ \\
\\
$^1$Waymo \quad $^2$Stanford University \quad $^3$Google Research\\

}
\vspace{-2em}
\twocolumn[{%
\renewcommand\twocolumn[1][]{#1}%
\maketitle
\vspace{-.8cm}
\input{figures/teaser}%
\vspace{-.3cm}
}]

\setlength{\abovedisplayskip}{0.05\abovedisplayskip}
\setlength{\belowdisplayskip}{0.05\belowdisplayskip}
\setlength{\abovecaptionskip}{0.05\abovecaptionskip}
\setlength{\belowcaptionskip}{0.05\belowcaptionskip}

\input{sections/0_abstract.tex}
\input{sections/1_introduction.tex}
\input{sections/2_related_work.tex}

\input{sections/3_method.tex}
\input{sections/4_experiments.tex}
\input{sections/5_conclusions.tex}

\input{sections/X_appendix}


{\small
\bibliographystyle{ieee_fullname}
\bibliography{egbib}
}

\end{document}

%% file: preamble.tex
\usepackage{enumitem}
\usepackage[numbers]{natbib}
\usepackage{colortbl}

\usepackage{multicol}

\renewcommand{\paragraph}[1]{\vspace{0em}\noindent\textbf{#1}.}

\usepackage{color}
\usepackage{xcolor}
\definecolor{turquoise}{cmyk}{0.65,0,0.1,0.3}
\definecolor{purple}{rgb}{0.65,0,0.65}
\definecolor{dark_green}{rgb}{0, 0.5, 0}
\definecolor{orange}{rgb}{0.8, 0.6, 0.2}
\definecolor{red}{rgb}{0.8, 0.2, 0.2}
\definecolor{darkred}{rgb}{0.6, 0.1, 0.05}
\definecolor{blueish}{rgb}{0.0, 0.3, .6}
\definecolor{light_gray}{rgb}{0.7, 0.7, .7}
\definecolor{pink}{rgb}{1, 0, 1}
\definecolor{greyblue}{rgb}{0.25, 0.25, 1}
\definecolor{tab_blue}{HTML}{1f77b4}
\definecolor{tab_orange}{HTML}{ff7f0e}
\definecolor{LightRed}{rgb}{0.99,0.89,0.89}

\definecolor{mesh_misty_rose}{HTML}{e6aaa3}
\definecolor{mesh_yellow}{HTML}{ffba00}

\renewcommand{\paragraph}[1]{\vspace{.2em}\noindent\textbf{#1}.}

\newcommand{\SupplementaryMaterial}{{\color{darkred} supplementary material}\xspace}

%% file: math_commands.tex
\usepackage{amsmath,amsfonts,bm}

\def\eqref#1{equation~\ref{#1}}

\def\1{\bm{1}}

\def\eps{{\epsilon}}

\def\rvc{{\mathbf{c}}}
\def\rvd{{\mathbf{d}}}

\def\rvo{{\mathbf{o}}}

\def\rvr{{\mathbf{r}}}
\def\rvs{{\mathbf{s}}}

\def\rvx{{\mathbf{x}}}

\def\rvz{{\mathbf{z}}}

\def\ervd{{\textnormal{d}}}

\def\rmC{{\mathbf{C}}}

\def\rmP{{\mathbf{P}}}

\DeclareMathAlphabet{\mathsfit}{\encodingdefault}{\sfdefault}{m}{sl}
\SetMathAlphabet{\mathsfit}{bold}{\encodingdefault}{\sfdefault}{bx}{n}

\def\gD{{\mathcal{D}}}
\def\gE{{\mathcal{E}}}

\def\gL{{\mathcal{L}}}

\def\gN{{\mathcal{N}}}

\def\gS{{\mathcal{S}}}

\def\sP{{\mathbb{P}}}

\newcommand{\E}{\mathbb{E}}

\DeclareMathOperator*{\argmin}{arg\,min}

%% file: figures/teaser.tex
\begin{center}
\captionsetup{type=figure}
\includegraphics[width=.85\linewidth]{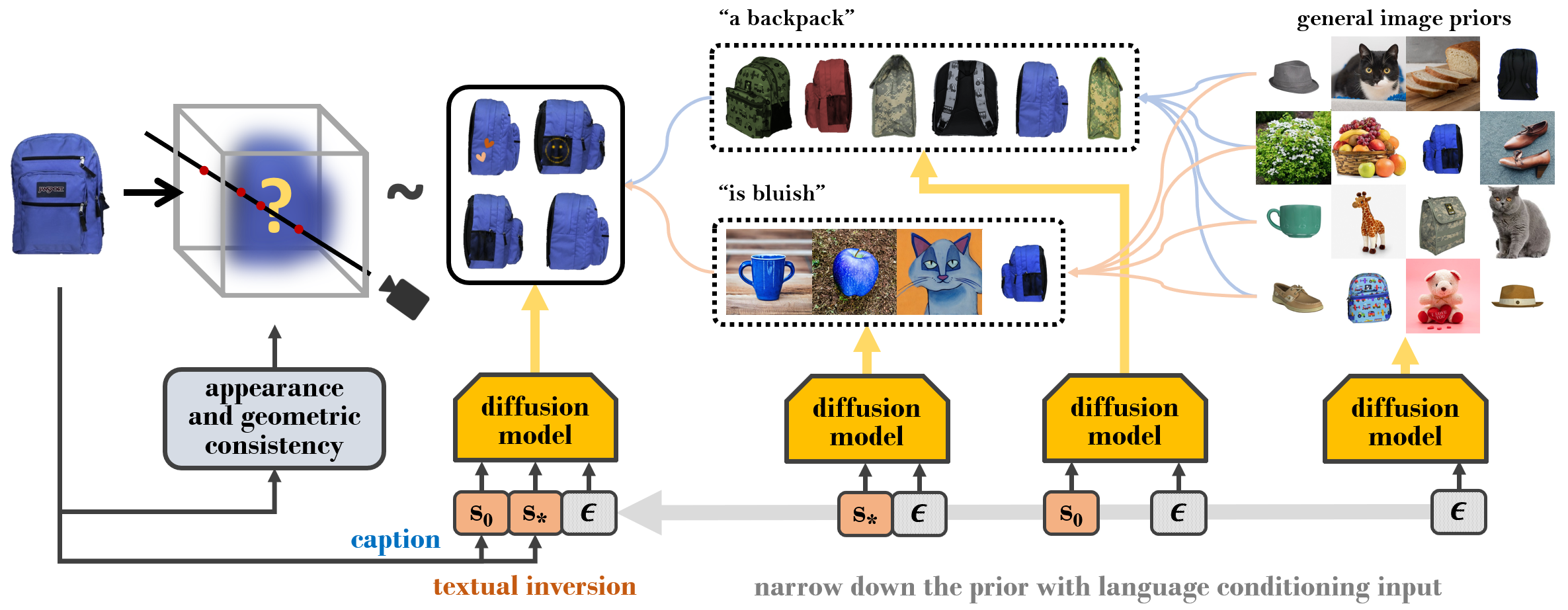}
\vspace{-.6em}
\captionof{figure}{
\textbf{From left to right:} We present a single-image NeRF synthesis framework for in-the-wild images without 3D supervision by leveraging general priors from large-scale image diffusion models. Given an input image, we optimize for a NeRF by minimizing an image distribution loss for arbitrary-view renderings with the diffusion model conditioned on the input image.
We design a two-section semantic feature as the conditioning input to the diffusion model.
The first section is the image caption $\rvs_0$  which carries the overall semantics; the second section is a text embedding $\rvs_*$ extracted from the input image with textual inversion, which captures additional visual cues.
Our two-section semantic feature provides an appropriate image prior, allowing the synthesis of a realistic NeRF coherent to the input image.
}
\vspace{1.3em}
\label{fig:teaser}
\end{center}

%% file: sections/0_abstract.tex
\begin{abstract}
\vspace{-1em}
2D-to-3D reconstruction is an ill-posed problem, yet humans are good at solving this problem due to their prior knowledge of the 3D world developed over years. Driven by this observation, we propose \textbf{NeRDi}, a single-view \textbf{NeR}F synthesis framework with general image priors from 2D \textbf{di}ffusion models. Formulating single-view reconstruction as an image-conditioned 3D generation problem, we optimize the NeRF representations by minimizing a diffusion loss on its arbitrary view renderings with a pretrained image diffusion model under the input-view constraint. We leverage off-the-shelf vision-language models and introduce a two-section language guidance as conditioning inputs to the diffusion model. 
This is essentially helpful for improving multiview content coherence as it narrows down the general image prior conditioned on the semantic and visual features of the single-view input image.
Additionally, we introduce a geometric loss based on estimated depth maps to regularize the underlying 3D geometry of the NeRF. 
Experimental results on the DTU MVS dataset show that our method can synthesize novel views with higher quality even compared to existing methods trained on this dataset. 
We also demonstrate our generalizability in zero-shot NeRF synthesis for in-the-wild images.

\let\thefootnote\relax\footnote{*Work done as an intern at Waymo.}

\end{abstract}

\vspace{-2em}

%% file: sections/1_introduction.tex
\vspace{-.5em}
\section{Introduction}
\label{sec:introduction}
\vspace{-.5em}

Novel view synthesis is a long-existing problem in computer vision and computer graphics.
Recent progresses in neural rendering such as NeRFs~\cite{mildenhall2021nerf} have made huge strides in novel view synthesis.
%
%
Given a set of multi-view images with known camera poses, NeRFs represent a static 3D scene as a radiance field parametrized by a neural network, which enables rendering at novel views with the learned network.
A line of work has been focusing on reducing the required inputs to NeRF reconstructions, ranging from dense inputs with calibrated camera poses to sparse images~\cite{jain2021putting,niemeyer2022regnerf,yu2021pixelnerf} with noisy or without camera poses~\cite{wang2021nerf}.
%
Yet the problem of NeRF synthesis from \emph{one single view} remains challenging due to its ill-posed nature,
as the one-to-one correspondence from a 2D image to a 3D scene does not exist.
Most existing works formulate this as a reconstruction problem and tackle it by training a network to predict the NeRF parameters from the input image \cite{yu2021pixelnerf,dupont2020equivariant}. 
But they require matched multiview images with calibrated camera poses as supervision, which is inaccessible in many cases such as images from the Internet or captured by non-expert users with mobile devices.
%
Recent attempts have been focused on relaxing this constraint by using unsupervised training with novel-view adversarial losses and self-consistency~\cite{mi2022im2nerf, ye2021shelf}. 
But they still require the test cases to follow the training distribution which limits their generalizability.
There is also work~\cite{vasudev2022ss3d} that aggregates priors learned on synthetic multi-view datasets and transfers them to in-the-wild images using data distillation. 
But they are missing fine details with poor generalizability to unseen categories.

Despite the difficulty of 2D-to-3D mapping for computers, it is actually not a difficult task for human beings.
Humans gain knowledge of the 3D world through daily observations and form a common sense of how things should look like and should not look like. 
Given a specific image, they can quickly narrow down their prior knowledge to the visual input. 
This makes humans good at solving ill-posed perception problems like single-view 3D reconstruction.
Inspired by this, we propose a single-image NeRF synthesis framework without 3D supervision by leveraging large-scale diffusion-based 2D image generation model (Figure \ref{fig:teaser}).
Given an input image, we optimize for a NeRF by minimizing an image distribution loss for arbitrary-view renderings with the diffusion model conditioned on the input image.
An unconstrained image diffusion is the `general prior' which is inclusive but also vague. 
To narrow down the prior knowledge and relate it to the input image, we design a two-section semantic feature as the conditioning input to the diffusion model. 
The first section is the image caption which carries the overall semantics; the second is a text embedding extracted from the input image with textual inversion \cite{gal2022image}, which captures additional visual cues. 
These two sections of language guidance facilitate our realistic NeRF synthesis with semantic and visual coherence between different views.
In addition, we introduce a geometric loss based on the estimated depth of the input view for regularizing the underlying 3D structure.
Learned with all the guidance and constraints, our model is able to leverage the general image prior and perform zero-shot NeRF synthesis on single image inputs. Experimental results show that we can generate high quality novel views from diverse in-the-wild images. To summarize, our key contributions are:

\begin{itemize}[leftmargin=*]
    \setlength{\itemsep}{1pt}
    \setlength{\parskip}{0pt}
    \setlength{\parsep}{0pt}
    \item We formulate single-view reconstruction as a conditioned 3D generation problem and propose a single-image NeRF synthesis framework without 3D supervision, using 2D  priors from diffusion models trained on large image datasets.
    \item We design a two-section semantic guidance to narrow down the general prior knowledge conditioned on the input image, enforcing synthesized novel views to be semantically and visually coherent.
    \item We introduce a geometric regularization term on estimated depth maps with 3D uncertainties.
    \item We validate our zero-shot novel view synthesis results on the DTU MVS \cite{jensen2014large} dataset, achieving higher quality than supervised baselines.
    We also demonstrate our capability of generating novel-view renderings with high visual quality on in-the-wild images.
\end{itemize}
\vspace{-1em}

%% file: sections/2_related_work.tex
\vspace{-.5em}
\section{Related Work}
\label{sec:related_work}
\vspace{-.5em}

\paragraph{Novel view synthesis with NeRF}
The recently proliferating NeRF representation \cite{mildenhall2021nerf} has shown great success in novel view synthesis, which is a long-existing task in computer graphics and vision.
Combining differentiable rendering \cite{li2018differentiable, zhang2019differential, zhang2020path, zhang2021path} with neural network scene parametrizations, NeRF is able to recover the underlying 3D scene from a collection of posed images and render it at novel views realistically.
A number of follow-up works have been focusing on relaxing NeRF inputs to less informative data such as unposed images \cite{yariv2020multiview, wang2021nerf, meng2021gnerf} or sparse views \cite{niemeyer2022regnerf, jain2021putting, roessle2022dense, deng2022depth}. As less data gives rise to a more complex optimization landscape, a variety of regularization losses have been studied, for example: RegNeRF \cite{niemeyer2022regnerf} regularizes the geometry and appearance of patches, DDP \cite{roessle2022dense} and DS-NeRF \cite{deng2022depth} regularize the depth maps, DietNeRF \cite{jain2021putting} enforces semantic consistency between views by minimizing a CLIP \cite{radford2021learning} feature loss, and GNeRF \cite{meng2021gnerf} adopts a patch-based adversarial loss.
Another line of work learns NeRF-based novel-view prediction for few- or single-image inputs by pre-training a scene prior on a large dataset of 3D scenes containing dense views \cite{chen2021mvsnerf,chibane2021stereo,liu2022neural,trevithick2021grf,wang2021ibrnet,yu2021pixelnerf}.
With additional self-supervision techniques such as equivariance \cite{dupont2020equivariant} or cycle-consistency \cite{mi2022im2nerf}, the learning of scene priors can be done simply from sparse- or single-view data, or even purely from unposed image collections with an image adversarial loss \cite{schwarz2020graf, niemeyer2021giraffe, chan2021pi, chan2022efficient}.
These two lines of works both have their specialties and constraints: the first is generalizable to any scene configurations, but is also less competitive in the more challenging scenarios such as single-image novel view synthesis with high quality requirements; the second, on the other hand, has strong ability of inferring unseen novel views from very limited inputs, but is also restricted to certain scene categories modeled by their scene priors learned from the training data.
In our work, we leverage a diffusion-based image prior for NeRF synthesis that is general enough for modeling variations of in-the-wild images while having the adaptivity to each specific input image.

\paragraph{Diffusion-based generative models}
Denoising diffusion probabilistic models \cite{ho2020denoising,song2020denoising}, or score-based generative models \cite{song2019generative, song2020score}, have recently caught a surge of interests due to their simple designs and excellent performances across a variety of computer vision tasks such as image generation \cite{song2019generative, ho2020denoising,song2020denoising, song2020score}, completion \cite{song2020score, saharia2022palette}, and editing \cite{meng2021sdedit, kim2022diffusionclip}. In visual content creation, language-guided image diffusion models such as DALL-E2 \cite{ramesh2022hierarchical}, Imagen \cite{saharia2022photorealistic} and Stable Diffusion \cite{Rombach_2022_CVPR} have shown great success in generating photo realistic images with strong semantic correlation to the given text-prompt inputs.
In additional to the success of 2D image diffusion models, more recent works have also extend diffusion models to 3D content generation.
\cite{luo2021diffusion, zhou20213d} generate 3D pointclouds with point diffusions.
3DiM \cite{watson2022novel} shows uncertainty-aware novel view synthesis with image diffusions conditioned on input views and poses, but it does not have guaranteed multiview consistency as no underlying 3D representation is adopted.
More related to ours are DreamFusion \cite{poole2022dreamfusion} and GAUDI \cite{bautista2022gaudi} that also generate NeRFs with diffusions:
\cite{poole2022dreamfusion} generates NeRFs under language guidance by optimizing for their renderings at randomly sampled views with a 2D image diffusion model \cite{saharia2022photorealistic}; \cite{bautista2022gaudi} trains a diffusion model on the latent space of NeRF scenes, but the learned scene distribution is limited to a set of indoor 3D scenes and does not generalize to in-the-wild images.
Similar to \cite{poole2022dreamfusion}, we also leverage 2D image diffusions to optimize for the NeRF renderings at novel views, but instead of unconstrained NeRF generation with user-specified language inputs, we study how to faithfully capture the the features of single-view image inputs and use it to constrain the novel-view image distributions.

%% file: sections/3_method.tex
\vspace{-.5em}
\section{Method}
\label{sec:method}
\vspace{-.5em}

\input{figures/method_overview}

An overview of our method is shown in Figure \ref{fig:method_overview}.
Given an input image $\rvx_0$, we would like to learn a NeRF representation $F_\omega: (x, y, z) \to (\rvc, \sigma)$ as its 3D reconstruction\footnote[2]{Here we use a Lambertian NeRF without view direction inputs for enforcing stronger multiview consistency.}.
The NeRF holds the rendering equation that, for any camera view with pose $\rmP$, one can sample camera rays $\rvr(t) = \rvo + t\rvd$ and render the image $\rvx$ at this view with
\begin{equation}
    \hat{\rmC}(\rvr) = \int_{t_n}^{t_f} T(t) \sigma(t) \rvc(t) \ervd t
\end{equation}
where $T(t) = \exp\left( -\int_{t_n}^t \sigma(s) \ervd s \right)$. For more details,  please refer to Mildenhall~\etal\cite{mildenhall2021nerf}.
For simplicity, we denote this whole rendering equation by $\rvx = f(\rmP, \omega)$ which means NeRF $f$ renders image $\rvx$ at camera pose $\rmP$ with parameters $\omega$.
Instead of predicting the NeRF parameters $\omega$ from $\rvx_0$ in a forward pass, we formulate this as a conditioned 3D generation problem
\begin{equation}
    f(\cdot,\omega) \sim \text{3D scene distribution} ~|~ f(\rmP_0,\omega) = \rvx_0
\end{equation}
where we optimize the NeRF to follow a 3D scene distribution conditioned on that its rendering $f(\rmP_0,\omega)$ at a given view $\rmP_0$ should be the input image $\rvx_0$

Directly learning the 3D scene distribution prior requires large 3D datasets, which is less straightforward to acquire and restricts its application to unseen scene categories.
To enable better generalizability to in-the-wild scenarios, we instead leverage 2D image priors and reformulate the objective into
\begin{equation}
    \forall \rmP, ~f(\rmP,\omega) \sim \sP~|~ f(\rmP_0,\omega) = \rvx_0
\end{equation}
where the optimization is conducted on images $f(\rmP, \omega)$ rendered at arbitrarily sampled views, pushing them to follow an image prior $\sP$ while satisfying the constraint $\rvx_0 = f(\rmP_0, \omega)$.
The overall objective can be written as maximizing the conditional probability
\begin{equation}
    \max_\omega \E_\rmP ~ \sP \left( f(\rmP,\omega) \,|\, f(\rmP_0,\omega) = \rvx_0,\, \rvs \right).
\end{equation}
Here, $\rvs$ is an additional semantic guidance term that we apply to further restrict the prior image distribution to fit the generation context. In contrast to DreamFusion \cite{poole2022dreamfusion} which also utilizes language-guided image diffusion model as 2D image priors for sampled views, our main contribution stands in our approach for further constraining the identity of the generated 3D volume to be consistent with the inputs.

We cover more details on this novel-view distribution loss in Sec. \ref{sec:method:novel_view}. We utilize natural language descriptions of the scene as the semantic guidance $\rvs$. More details on this will be discussed in Sec. \ref{sec:method:semantics}. In addition, as the image diffusion model only operates on the rendered rgb colors, we further apply a geometric regularization with a depth map estimated at the input view to facilitate the NeRF optimization (Sec. \ref{sec:method:depth})

\vspace{-.5em}
\subsection{Novel View Distribution Loss}
\label{sec:method:novel_view}
\vspace{-.5em}

Denoising Diffusion Probabilistic Models (DDPM) are a type of generative models that learn a distribution over training data samples.
Recently, there are many advances in language guided image synthesis with diffusion models.
We build our method upon the recent Latent Diffusion Model (LDM) \cite{Rombach_2022_CVPR} for its high quality and efficiency in image generation. It adopts a pre-trained image auto-encoder with an encoder $\gE(\rvx) = \rvz$ mapping images $\rvx$ into latent codes $\rvs$ and a decoder $\gD(\gE(\rvx)) = \rvx$ recovering the images.
The diffusion process is then trained in the latent space by minimizing the objective
\begin{equation}
\label{eq:latent_diffusion}
    \E_{\rvz\sim\gE(\rvx), \rvs, \eps\sim\gN(0, 1), t}
    \left[ \| \eps - \eps_\theta(\rvz_t, t, c_\theta(\rvs)) \|^2_2 \right].
\end{equation}
where $t$ is a diffusion time scale, $\eps\sim\gN(0, 1)$ is a random noise sample, $\rvz_t$ is the latent code $\rvz$ noised to time $t$ with $\eps$, and $\eps_\theta$ is the denoising network with parameters $\theta$ to regress the noise $\eps$.
The diffusion model also takes a conditioning input $\rvs$ which is encoded as $c_\theta(\rvs)$ and serves as guidance in the denoising process. For text-to-image generation models such as the LDM, $c_\theta$ is a pre-trained large language model that encodes the conditional text $\rvs$.

In a pre-trained diffusion model, the network parameters $\theta$ are fixed, and we can instead optimize for the input image $\rvx$ with the same objective which transforms $\rvx$ to follow the image distribution priors conditioned on $\rvs$.
Let $\rvx = f(\rmP, \omega)$ be our NeRF rendering at arbitrarily sampled view $\rmP$, we can back propagate gradients to the NeRF parameters $\omega$ and thus get a stochastic gradient descent on $\omega$.

\vspace{-.5em}
\subsection{Semantics-Conditioned Image Priors}
\label{sec:method:semantics}
\vspace{-.5em}

\input{figures/method_text_guidance}

We argue that the prior distribution over all in-the-wild images is not specific enough to guide the novel view synthesis from an arbitrary image.
We thus introduce a well-designed guidance $\rvs$ that narrows down the generic prior over natural images to a prior of images related to the input image $\rvx_0$. 
Here we choose text as the guidance, which is flexible for describing arbitrary input images. 
Text-to-image diffusion models such as LDM utilize a pre-trained large language model as the language encoder to learn a conditional distribution over images conditioned on language. 
This serves as a natural gateway for us to utilize language as a means to restrict the image prior space.

The most straightforward way of getting a text prompt from the input image is to use an image captioning or classification network $\gS$ trained on (image, text) datasets and predict a text $\rvs_0 = \gS(\rvx_0)$.
However, while text description can summarize the semantics of the image, it leaves a huge space of ambiguities, making it hard to include all the visual details in the image especially with limited prompt length.
In Figure \ref{fig:method_text_guidance} top row, we show the images generated with the caption ``a collection of products'' from the input image on the left. While their semantics are highly accurate with respect to the language description, the generated images have very high variances in their visual patterns and low correlations to the input image.

Textual inversion \cite{gal2022image}, on the other hand, optimizes for the text embedding of one or few images from a text-based image diffusion model.
With the LDM Equation \ref{eq:latent_diffusion}, we can optimize for the text embedding $\rvs_*$ for the input image $\rvx_0$ by
\begin{equation}
    \rvs_* = \argmin_\rvs
    \E_{\rvz\sim\gE(\rvx_0), \rvs, \eps\sim\gN(0, 1), t}
    \left[ \| \eps - \eps_\theta(\rvz_t, t, c_\theta(\rvs)) \|^2_2 \right]
\end{equation}
In Figure \ref{fig:method_text_guidance} middle row, images generated with textual inversion are shown. The colors and visual cues of the input image are well captured (orange-colored elements, food, and even the brand logos). However, the semantics at the macro level is sometimes wrong (second column is a person playing sports). One reason is that, different from the multi-image scenarios where textual inversion can discover the common contents of these images, it is unclear for one single image what the key features are that the text embedding should focus on.

To reflect both semantic and visual characteristics of the input image in the novel view synthesis task, we combine these two methods by concatenating their text embeddings to form a joint feature $\rvs = [\rvs_0, \rvs_*]$ and use it as the guidance in the diffusion process in Equation \ref{eq:latent_diffusion}.
Figure \ref{fig:method_text_guidance} bottom row shows the images generated with this joint feature, with balanced semantics and visual cues.

\vspace{-.5em}
\subsection{Geometric Regularization}
\label{sec:method:depth}
\vspace{-.5em}

\input{figures/method_depth_ambiguity.tex}

While image diffusion shapes the appearance of the NeRF, multiview consistency is difficult to enforce as the underlying 3D geometry can be different even with the same image rendering \cite{lehar2003world, mitra2009shadow}, making the gradient back-propagation (from the image diffusion to the NeRF parameters $\omega$) highly non-controllable.
To this end, we further incorporate a geometric regularization term on the input view depth to alleviate this issue.
We adopt the Dense Prediction Transformer (DPT) model \cite{ranftl2021vision} trained on 1.4 million images for zero-shot monocular depth estimation and apply it to the input image $\rvx_0$ to estimate a depth map $\rvd_{0,\text{est}}$.
We use this estimated depth to regularize the depth
\begin{equation}
    \hat{\rvd}_0 = \int_{t_n}^{t_f} \sigma(t) \ervd t.
\end{equation}
rendered by the NeRF at input view $\rmP_0$.
Due to the ambiguities of the estimated depth (including scales, shifts, camera intrinsics) and estimation error (Figure \ref{fig:method_depth_ambiguity}), we cannot back project pixels with depth to 3D and compute the regularization directly. Instead, we maximize the \textit{Pearson} correlation between the estimated depth map and the NeRF-rendered depth
\begin{equation}
    \rho\left( \hat{\rvd}_0, \rvd_{0, \text{est}} \right) =
    \frac{\text{Cov}(\hat{\rvd}_0, \rvd_{0, \text{est}})}
    {\sqrt{\text{Var}(\hat{\rvd}_0) \text{Var}(\rvd_{0, \text{est}})}}
\end{equation}
which measures if the rendered depth distribution and the noisy estimated depth distribution are linearly correlated.

%% file: figures/method_overview.tex
\begin{figure}[t]
\centering
\includegraphics[width=.75\linewidth]{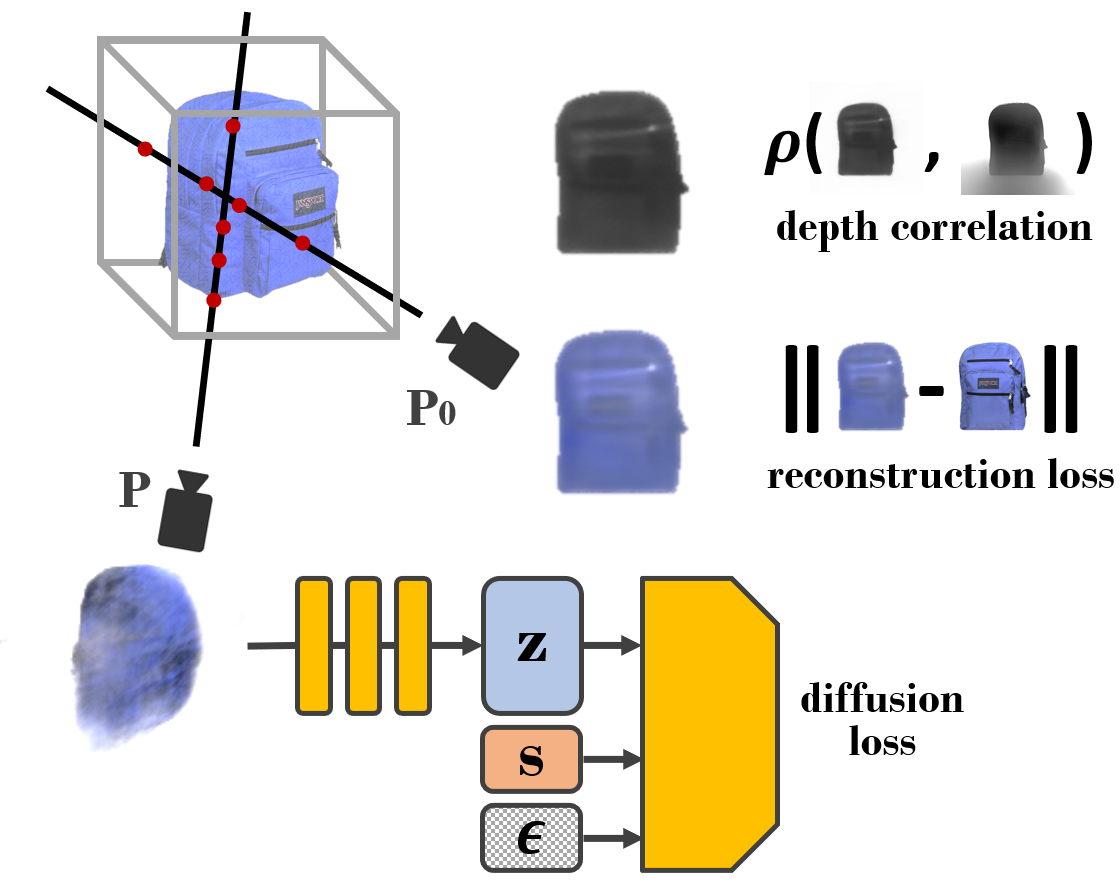}
\caption{\textbf{Method overview.}
We represent the underlying 3D scene as a NeRF and optimize for its parameters with three losses: a reconstruction loss at the fixed input view; a diffusion loss at arbitrarily sampled views which also takes a conditioning text input generated from the input image with our two-section feature extraction; and finally, a depth correlation loss at the input view regularizing the 3D geometry.
}
\vspace{-1.2em}
\label{fig:method_overview}
\end{figure}

%% file: figures/method_text_guidance.tex
\begin{figure*}[t]
\centering
\includegraphics[width=.8\linewidth]{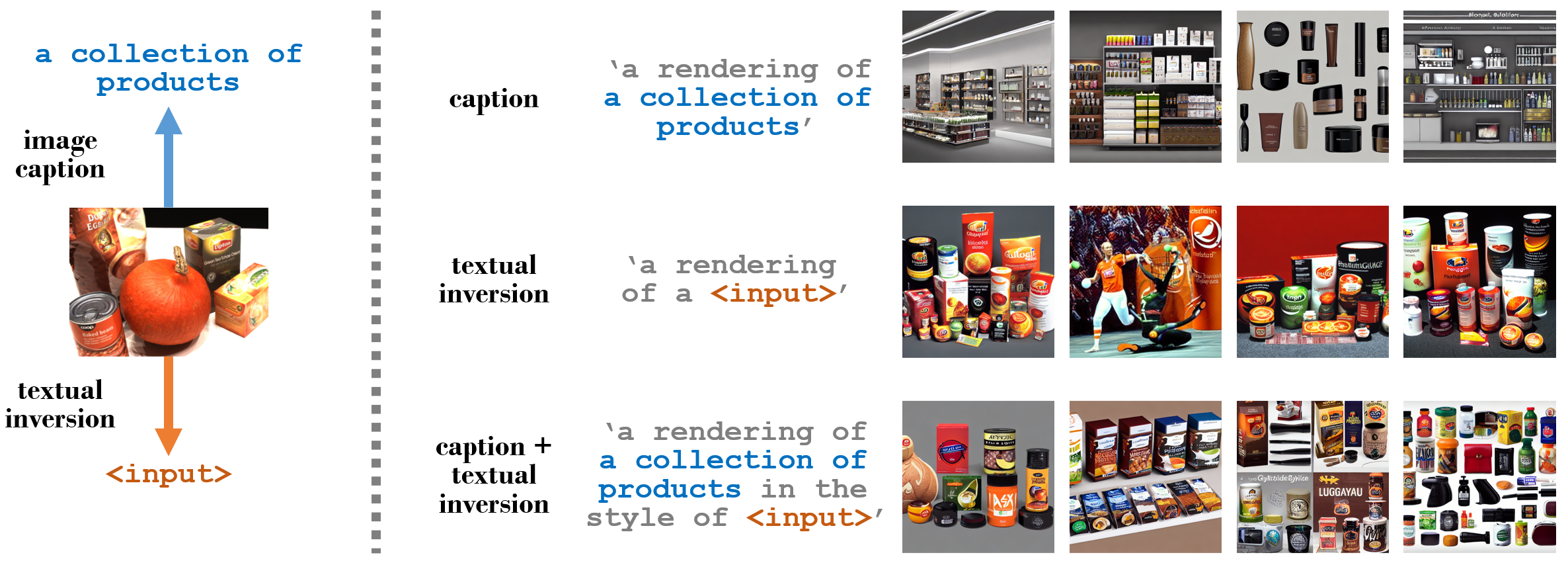}
\caption{\textbf{Image generation with different semantic guidance.}
\textbf{Top row:} Images generated with caption \textbf{\color{tab_blue}``a collection of products''}. The images follows the semantics well, but their content are of very high variance (can be any kind of products).
\textbf{Middle row:} Images generated purely with the latent embedding from \textbf{\color{tab_orange} textual inversion}. The color distribution and visual cues of the input image are well captured, but the semantics is not preserved (second column, the image is a person playing sports).
\textbf{Bottom row:} Images generated with combined image caption and textual inversion. Both semantic and visual features of the input image are addressed.
}
\label{fig:method_text_guidance}
\vspace{-1em}
\end{figure*}

%% file: figures/method_depth_ambiguity.tex
\begin{figure}[t]
\centering
\includegraphics[width=.6\linewidth]{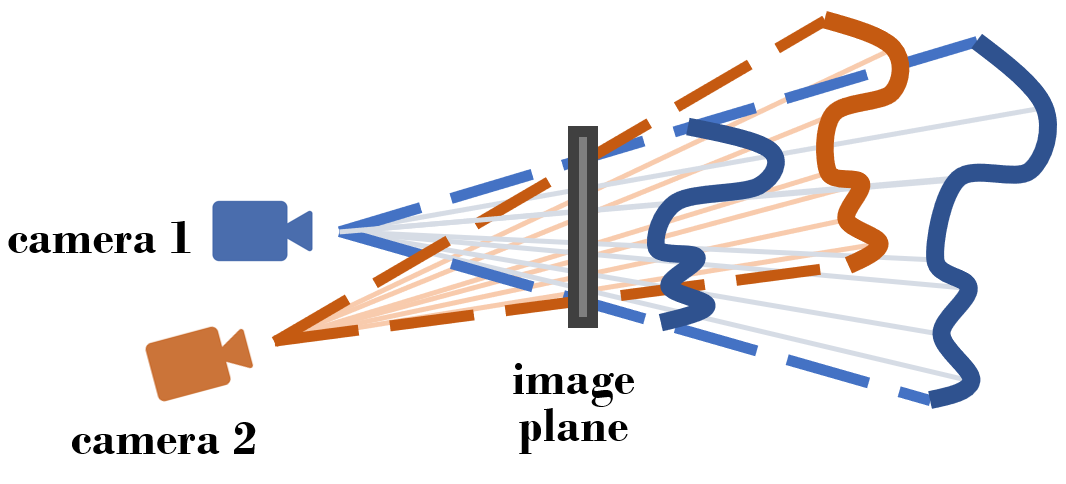}
\caption{\textbf{Ambiguity in estimated depth map.}
\vspace{-.5em}
}
\label{fig:method_depth_ambiguity}
\end{figure}

%% file: sections/4_experiments.tex
\vspace{-.5em}
\section{Experiments}
\label{sec:experiments}
\vspace{-.5em}


Now we demonstrate our efficacy in synthesizing realistic NeRFs with single-view inputs.
Section \ref{sec:exp:synthetic} presents a quantitative comparison between our method and the state-of-the-art single-view NeRF reconstruction methods on a synthetic dataset.
Section \ref{sec:exp:wild} shows a qualitative comparison as well as more synthesis results of our method on in-the-wild images.

\vspace{-.5em}
\subsection{Synthetic Scenes}
\label{sec:exp:synthetic}
\vspace{-.5em}

\input{figures/results_dtu_render.tex}
\input{tables/results_dtu.tex}

\paragraph{Setup}
We evaluate our method on the DTU MVS dataset \cite{jensen2014large} with 15 test scenes as specified in \cite{yu2021pixelnerf}.
For each input image, we use GPT-2 \cite{radford2019language} to generate a caption. We manually correct the obvious mistakes made by GPT-2 while trying our best to avoid introducing additional details. The scenes and their captions are listed in the \SupplementaryMaterial.

\paragraph{Implementation details}
For the NeRF model, we implement the multi-resolution grid sampler as described in \cite{muller2022instant}.
For the diffusion model, we employ the text-guided diffusion model from \cite{Rombach_2022_CVPR} which was pre-trained on the LAION-400M dataset \cite{schuhmann2021laion}.
While \cite{Rombach_2022_CVPR} operates on $512\times512$ images, NeRF's volumetric rendering at this resolution would incur an extensive computational burden. Thus, at the randomly sampled novel views, we render $128\times128$ images and resize them to $512\times512$ before feeding them to the encoder of \cite{Rombach_2022_CVPR}.
At the input view, we render at the same resolution as the input image to compute the image reconstruction and depth correlation losses.

\paragraph{Baselines}
We compare with two state-of-the-art single-view NeRF reconstruction algorithms, PixelNeRF \cite{yu2021pixelnerf} and its fine-tuned model with CLIP \cite{radford2021learning} feature consistency loss as proposed by DietNeRF \cite{jain2021putting}, both of which trained on the training set data from the DTU MVS dataset.
To gain better convergence, we use the predictions from \cite{yu2021pixelnerf} as an initialization for our 3D scene optimization. But our method is directly applied to the test scenes without any additional fine-tuning on the DTU training set.

\paragraph{Results}
Table \ref{tab:results_dtu} shows the quantitative comparison between our method and the baselines.
Following the convention, we report the standard image quality metrics PSNR and SSIM \cite{wang2004image}.
Our PSNR and SSIM are slightly lower than pixelNeRF \cite{yu2021pixelnerf} which directly learns the scene distributions from the DTU training set and are on par with DietPixelNeRF \cite{jain2021putting} which enforces semantic consistency between views.
However, we emphasize that these two metrics are less indicative in our scenario as they are local pixel-aligned similarity metrics between the synthesized novel views and the ground truth images but uncertainties naturally exist in single-view 3D inference.
The middle column of the first scene in Figure \ref{fig:results_dtu_render} shows an example of such uncertainty. The height of the tallest snack bag in the input image cannot be inferred as its top extrudes beyond the camera view. The width of the toy pig in the left column of the third scene is another example which cannot be inferred from the input side view. In both cases our method guesses its  novel view (bottom row) in a reasonable sense but different from the ground truth (top row).
In addition, we also measure novel views with LPIPS \cite{zhang2018unreasonable}, which is a perceptual metric computing the Mean Squared Error (MSE) between normalized features from all layers of a pre-trained VGG encoder \cite{simonyan2014very}.
Our method shows a significant improvement on this metric compared to the baselines as the diffusion model helps to improve image qualities while the language guidance maintains the multi-view semantic consistency.

Figure \ref{fig:results_dtu_render} shows a qualitative comparison between our method and the baselines.
With the scene initialization from \cite{yu2021pixelnerf}, our method removes the noises and blurriness, synthesizing high quality novel views.

\vspace{-.5em}
\subsection{Images in the Wild}
\label{sec:exp:wild}
\vspace{-.5em}

\input{figures/results_ganzebo}

\paragraph{Qualitative comparisons}
Figure \ref{fig:results_ganzebo} shows a qualitative comparison between our method and existing state-of-the-art single-image to 3D synthesis methods for in-the-wild images \cite{jain2021putting, vasudev2022ss3d}.
Input images are adopted from the Google Scanned Objects dataset \cite{downs2022google} with their category labels (`bag' and `hat') as captions.
Similar to ours, DietNeRF \cite{jain2021putting} uses an input-view constrained NeRF optimization technique where they minimize the CLIP \cite{radford2021learning} feature between arbitrary view renderings. While CLIP features enforce consistent appearances, they fail to capture the global semantics of the object.
SS3D \cite{vasudev2022ss3d} is a forward-prediction model for 3D geometries that transfers the priors learned on synthetic datasets to in-the-wild images with knowledge distillation.
While it generates more structured global geometries, it fails to capture the fine geometric details of the input image.
The geometries of the hats in the bottom rows are also incorrect, with only the silhouette shape preserved but the structure of ‘hat’ shape missing.

\input{figures/results_coco}

\paragraph{More results}
Figure \ref{fig:results_internet} shows our results on images of objects from the internet. The text prompts are words or phrases used to search for the images.
The backgrounds are masked out using an off-the-shelf dichotomous image segmentation network from \cite{qin2022}.
For each input, we show 3 different novel views that are distant from the input view.
Figure \ref{fig:results_coco} shows our results on images with more complex contents and backgrounds from the COCO dataset \cite{lin2014microsoft} which contains (image, caption) pairs. Within camera views close to the input, our model is still able to generate realistic renderings. But it can hardly generalize to distant views due to the limited capacity of the NeRF scene box.

\vspace{-.5em}
\subsection{Ablation Studies}
\vspace{-.5em}

\input{figures/results_ablation_text.tex}

We conduct ablation studies to show the efficacy of our two-section semantic guidance and geometric regularization.

\paragraph{Semantic guidance}
Figure \ref{fig:results_ablation_text_backpack} shows the ablation of the two text embeddings $\rvs_0$ from image captions and $\rvs_*$ from textual inversion.
Without the captions $\rvs_0$, the model fails to learn the overall semantics and cannot generate a meaningful object.
While both the full model and the caption-only one (without textual inversion) successfully generate backack novel views, the results without textual inversion $\rvs_*$ have more blurriness and noises. A zoom-in comparison is shown in Figure \ref{fig:results_ablation_text_zoomin}.

Figure \ref{fig:results_ablation_text_can} shows another comparison of models with and without textual inversion $\rvs_*$ on the can example from Figure \ref{fig:results_coco} left.
In the object regions visible to the input view, the full model better recovers the fine details (the white letters on the lateral); and in the invisible regions, the full model completes the appearances with coherent styles of the input (red and white textures at the back of the can), while the model without textual inversion does not have such appearance coherency.
The model with textual inversion can even synthesize the pull tab at the top (second column of the zoom-in views) by inferring from the input side view that this is a can containing drinks.

\input{figures/results_ablation_depth.tex}

\paragraph{Geometric regularization}
Figure \ref{fig:results_ablation_depth} shows an ablation on the geometric regularization term. Both image renderings and depth maps are visualized.
The full model is able to synthesize realistic novel views with coherent 3D geometry.
The model without the regularization on the input view depth can still generate realistic appearances at novel views with the diffusion model, but the underlying 3D geometry is erroneous and multi-view consistency is not enforced.
As a sanity check, we also visualize the results with only the depth loss but without the diffusion model. The model is unable to generate a realistic NeRF due to the 3D ambiguities of monocular depth as stated in Section \ref{sec:method:depth}.

%% file: figures/results_dtu_render.tex
\begin{figure*}[t]
\centering
\includegraphics[width=.9\linewidth]{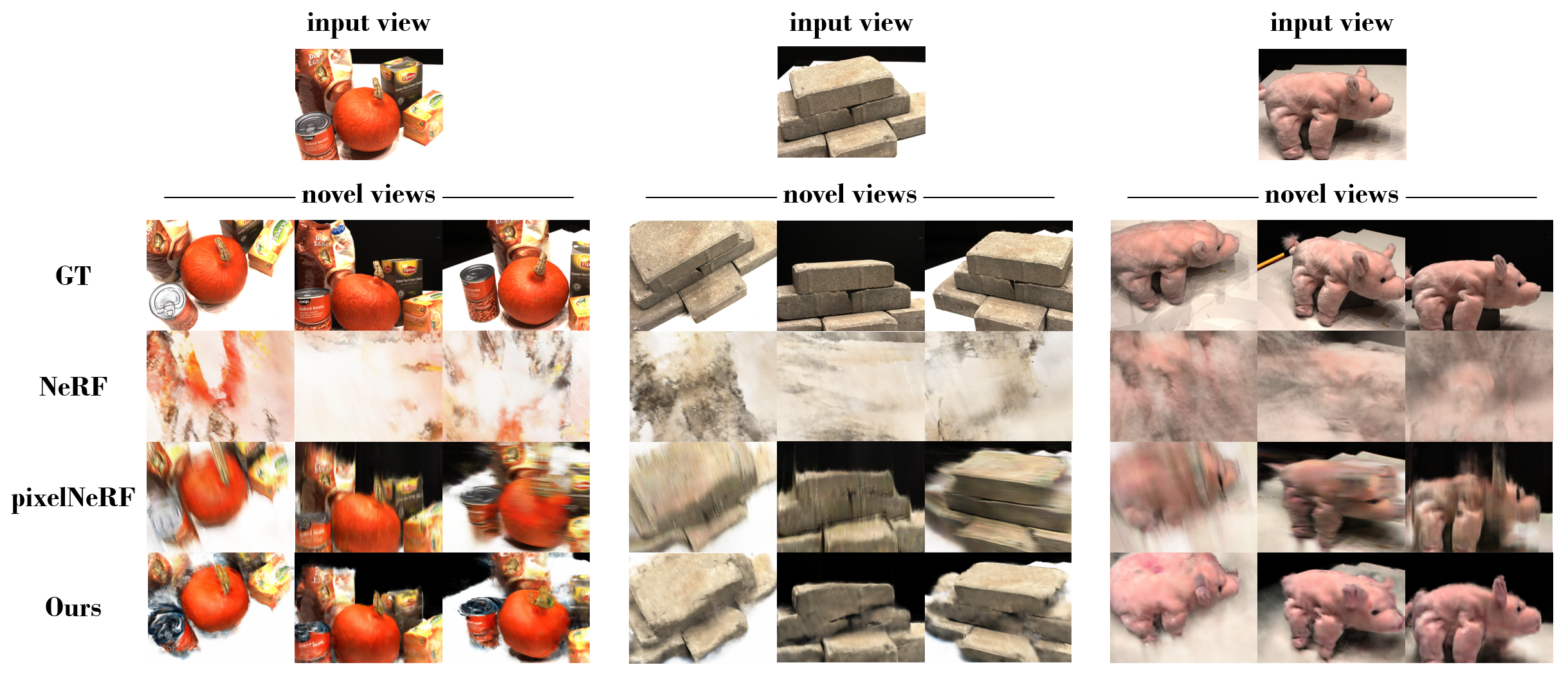}
\caption{\textbf{Single-image novel view synthesis results on the DTU test scenes.}
Vanilla NeRF cannot recover scenes from single image inputs due to the ill-posed nature of this problem.
While pixelNeRF can infer the novel view images with the prior from the DTU training set of similar scenes, its synthesized renderings remain noisy and blurry.
With a pixelNeRF initialization, our method is able to synthesize cleaner novel views with realistic geometries and appearances, despite having never been trained on this dataset.
\textbf{Uncertainties in novel view inference:} (The first scene, middle column) the exact height of the tallest snack bag cannot be inferred as the top goes outside of the camera view. (The third scene, left column) the width of the toy pig from the top view is undecidable from the input view. In both cases, our method guesses a reasonable answer in the synthesized novel view that is different from the ground truth. 
}
\vspace{-.7em}
\label{fig:results_dtu_render}
\end{figure*}

%% file: tables/results_dtu.tex
\newcolumntype{R}{>{\columncolor{LightRed}}c}
\begin{table}[t]
\caption{Single-image novel view synthesis results on DTU.}
\label{tab:results_dtu}
\centering
\begin{tabular}{lccR}
\toprule
\textbf{Method}                  & \textbf{PSNR} $\uparrow$  & \textbf{SSIM} $\uparrow$ & \textbf{LPIPS} $\downarrow$ \\ \midrule
NeRF                             & 8.000           & 0.286          & 0.703          \\
pixelNeRF                        & 15.550          & 0.537          & 0.535          \\
pixelNeRF, $\gL_{\text{MSE}}$ ft & \textbf{16.048} & \textbf{0.564} & 0.515          \\
DietPixelNeRF                    & 14.242          & 0.481          & 0.487          \\
Ours                             & 14.472          & 0.465          & \textbf{0.421} \\ \bottomrule
\end{tabular}
\vspace{-.5cm}
\end{table}

%% file: figures/results_ganzebo.tex
\begin{figure*}[t]
\centering
\includegraphics[width=.83\linewidth]{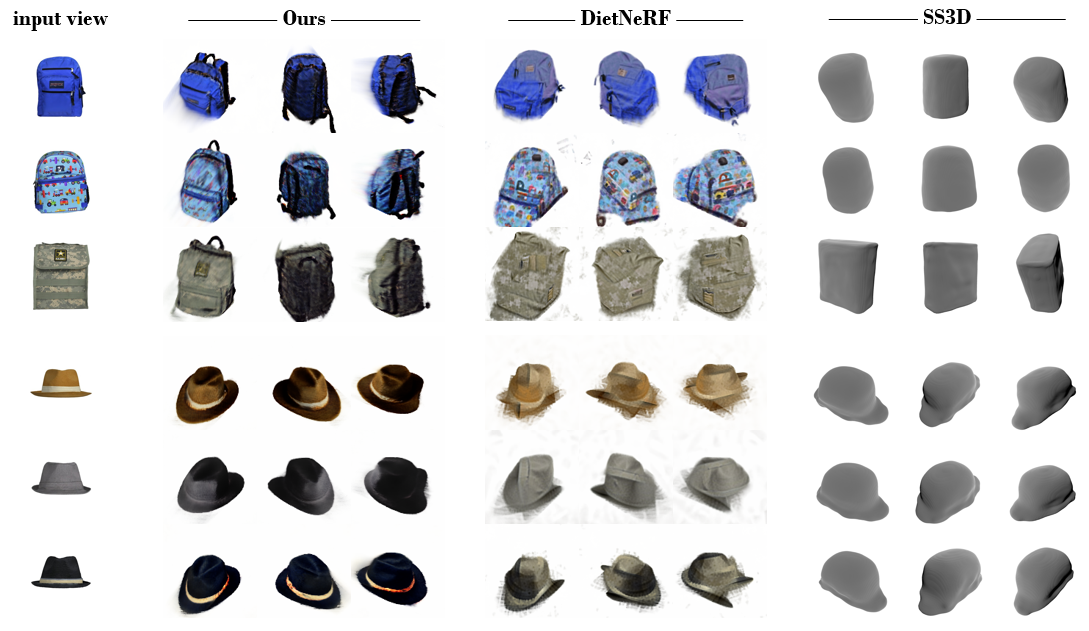}
\caption{\textbf{Novel view synthesis results on objects from the Google Scanned Objects Dataset.}
\textbf{Left:} Our results generated from single-word text inputs `backpack' (top 3 rows) and `hat' (bottom 3 rows).
\textbf{Middle:} DietNeRF \cite{jain2021putting} minimizes the CLIP feature distances between the input view and arbitrarily sampled views. This results in novel view renders with consistent textures and styles, but fails to capture the global semantic meaning. \emph{For a fair comparison, DietNeRF is also optimized with depth regularization.}
\textbf{Right:} SS3D \cite{vasudev2022ss3d} predicts coarse geometries in a consistency manner, but it fails to recover all the fine geometric details. Additionally, the geometries of the hats in the bottom rows are incorrect, with only the silhouette shape preserved but the structure of `hat' shape missing.
}
\label{fig:results_ganzebo}
\vspace{-1em}
\end{figure*}

%% file: figures/results_coco.tex
\begin{figure*}[t]
\centering
\begin{subfigure}[b]{.48\linewidth}
    \flushleft
    \includegraphics[height=5cm]{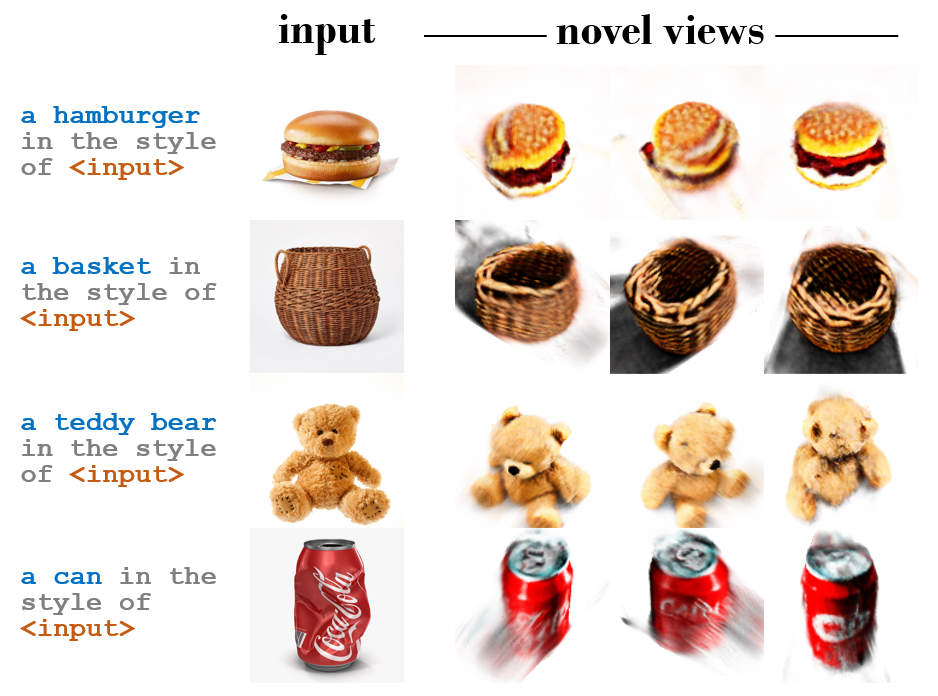}
    \caption{\textbf{Results on object-centric images from the internet with single-word or short phrase captions.}
    Input backgrounds are removed with \cite{qin2022}.
    }
    \label{fig:results_internet}
\end{subfigure}
~~
\begin{subfigure}[b]{.5\linewidth}
    \centering
    \includegraphics[height=5cm]{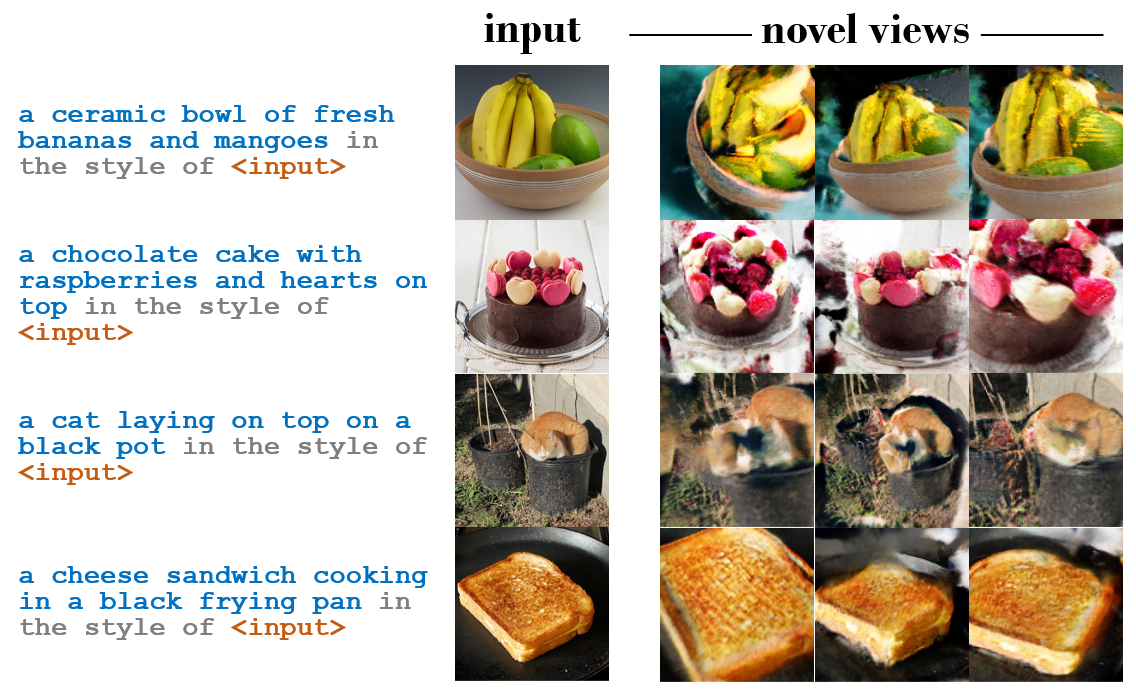}
    \caption{\textbf{Results on images from the COCO dataset \cite{lin2014microsoft}.}
    Input images have more complex contents with backgrounds and the captions are sentences.
    }
    \label{fig:results_coco}
\end{subfigure}
\caption{\textbf{Results on images in the wild.}}
\label{fig:results_wild}
\vspace{-1em}
\end{figure*}

%% file: figures/results_ablation_text.tex
\begin{figure}[t]
\centering
\begin{subfigure}[b]{\linewidth}
    \centering
    \includegraphics[width=.9\linewidth]{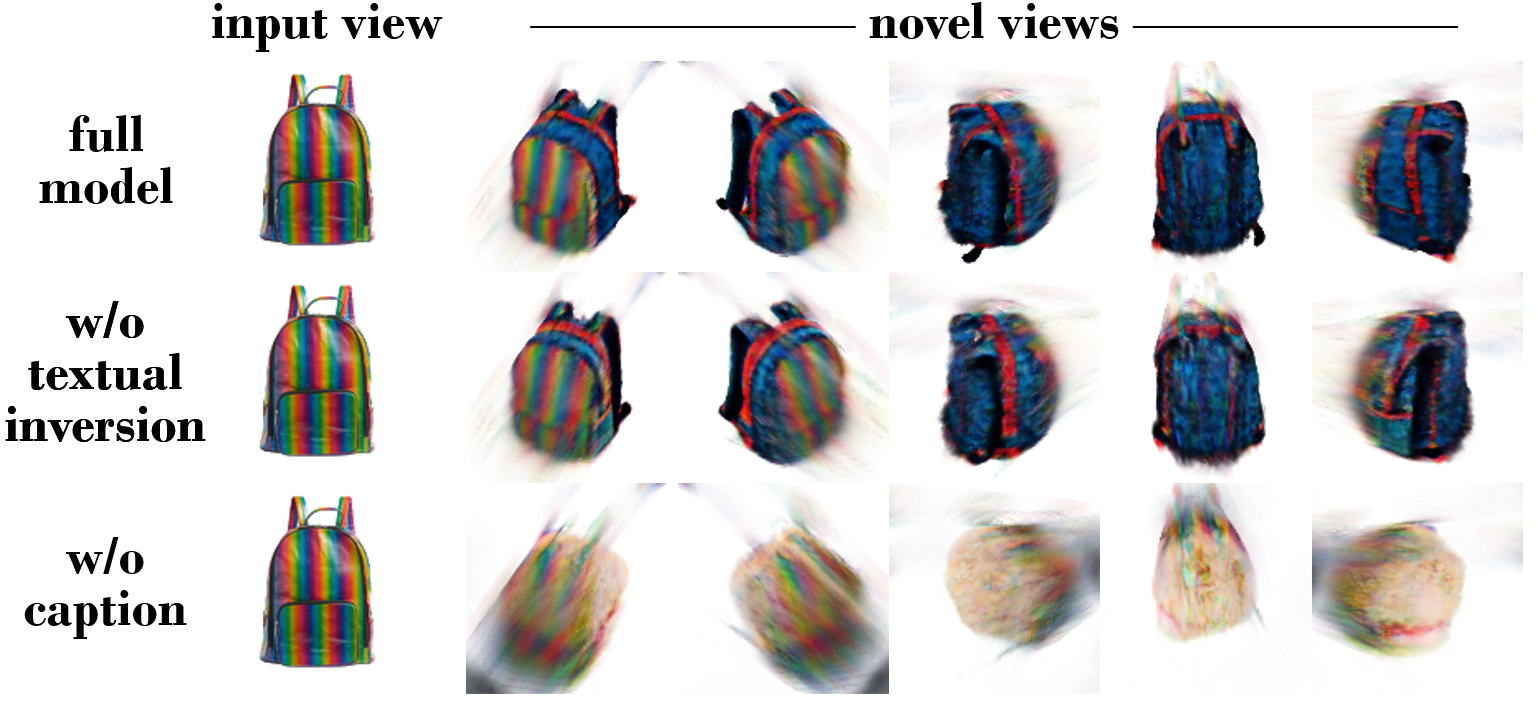}
    \caption{
    \textbf{Top row:} Full model.
    \textbf{Middle row:} Caption-only guidance without textual inversion. The model is still able to generate a shape strictly following the semantics and the input view appearance and geometric constraints, but struggles more in synthesizing the details. A zoom-in comparison is shown in \ref{fig:results_ablation_text_zoomin} below.
    \textbf{Bottom row:} Textual-inversion-only without caption. Textual inversion fails to capture the global semantics.
     \label{fig:results_ablation_text_backpack}
    }
\end{subfigure}
\begin{subfigure}[b]{\linewidth}
    \centering
    \includegraphics[width=.8\linewidth]{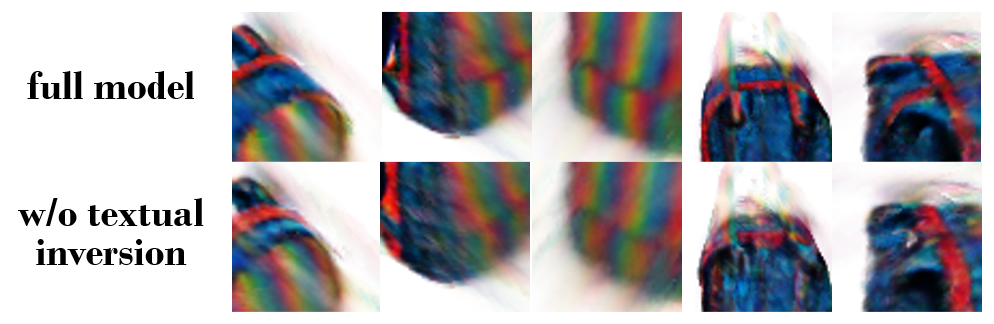}
    \caption{\textbf{A zoom-in comparison between full model results and results without textual inversion.}
    The full model shows better capability of synthesizing less blurry details.
    }
    \label{fig:results_ablation_text_zoomin}
\end{subfigure}
\begin{subfigure}[b]{\linewidth}
    \centering
    \includegraphics[width=.9\linewidth]{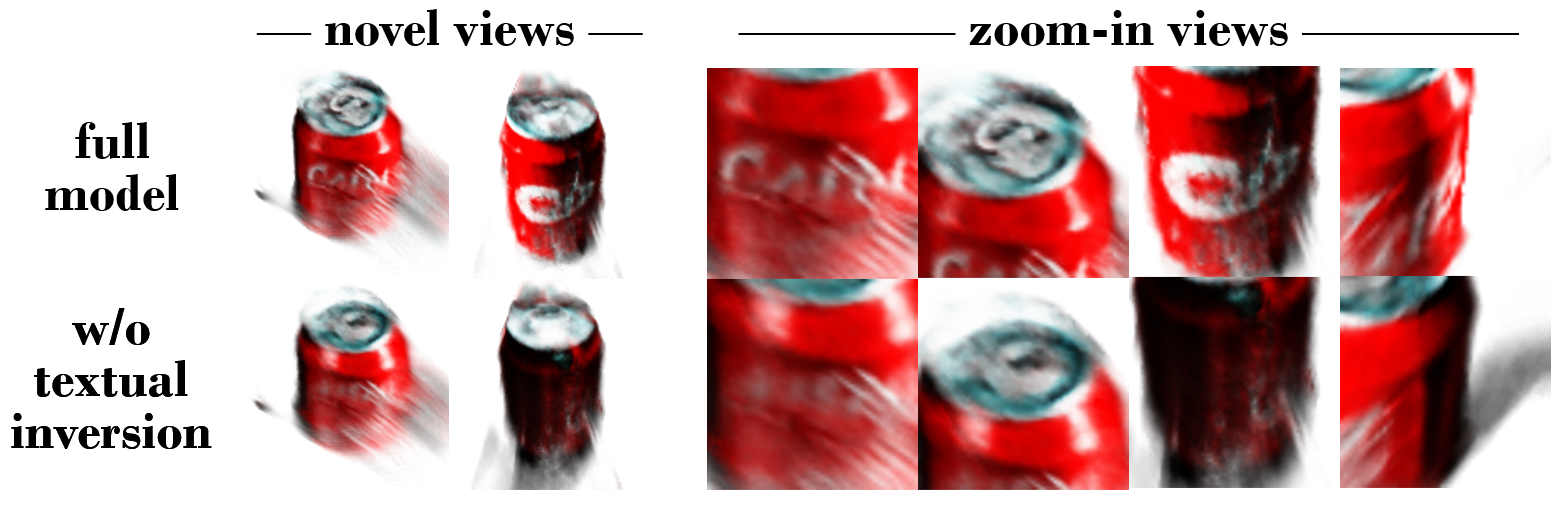}
    \caption{\textbf{Another comparison between models with ant without textual inversion.}
    The input is from \ref{fig:results_coco} left, bottom row.
    The full model is able to synthesize better texture details at visible regions as well as completing the invisible regions with similar textures, while the caption-only model renderings are more blurry and cannot fill in the invisible regions.
    }
     \label{fig:results_ablation_text_can}
\end{subfigure}
\caption{\textbf{Ablations on the two-section semantic guidance.}
}
\vspace{-.5em}
\label{fig:results_ablation_text}
\end{figure}

%% file: figures/results_ablation_depth.tex
\begin{figure}[t]
\centering
\includegraphics[width=.9\linewidth]{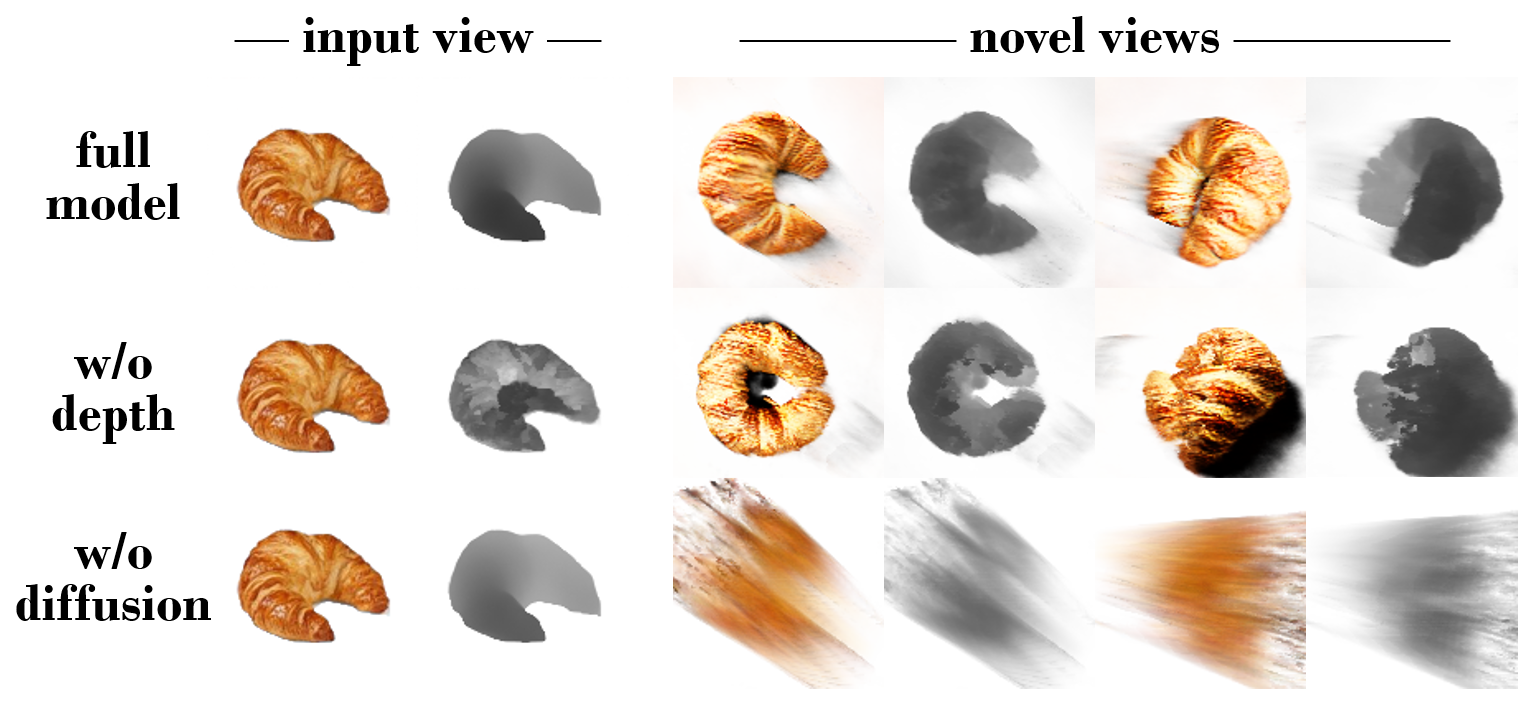}
\caption{\textbf{Ablations on the geometric regularization.}
Visualization of input view reconstruction and novel views on rendered images and depth maps.
\textbf{Top row:} The full model is able to synthesize realistic novel views while preserving geometric coherency.
\textbf{Middle row:} Without the depth correlation loss, the diffusion model is still able to generate reasonable appearances, but the underlying 3D geometry is erroneous and the novel views are inconsistent to the input view.
\textbf{Bottom row:} The input-view depth estimation cannot guide novel view synthesis by itself without the diffusion model due to 3D ambiguities.
}
\label{fig:results_ablation_depth}
\vspace{-1em}
\end{figure}

%% file: sections/5_conclusions.tex
\vspace{-.5em}
\section{Conclusions}
\label{sec:conclusions}
\vspace{-.5em}

\input{figures/limitations}

In this paper, we propose a novel framework for zero-shot single-view NeRF synthesis for images in the wild without 3D supervision.
We leverage the general image priors in 2D diffusion models and apply them to the 3D NeRF generation conditioned on the input image.
To efficiently use these priors in synthesizing consistent views, we design a two-section language guidance as conditioning inputs to the diffusion model which unifies the semantic and visual features of the input image.
To our knowledge, we are the first to combine semantic and visual features in the text embedding space and apply it to novel view synthesis.
In addition, we introduce a geometric regularization term while addressing the 3D ambiguity of monocular-estimated depth maps.
Our experimental results show that, with well-designed guidance and constraints, one can leverage general image priors to specific image-to-3D, enabling us to build generalizable and adaptable reconstruction frameworks.

\paragraph{Limitations and future work}
As our method relies on multiple large pre-trained image models \cite{Rombach_2022_CVPR, ranftl2021vision, qin2022,radford2019language}, any biases in these models will affect our synthesis results.
Figure \ref{fig:limitation_bias} shows an example where the image diffusion model \cite{Rombach_2022_CVPR} can generate two shoes even the text prompt is ``a single shoe'', resulting in our synthesized NeRF showing the features of multiple shoes. 
Our method is also less robust to highly deformable instances, as our language guidance focuses on semantics and styles but lacks a global description of physical states and dynamics. Figure \ref{fig:limitation_deformable} shows such a failure case. Renderings from each independent view are visually plausible but represent different states of the same instances.

Besides, while formulation-wise the optimization is applicable to any scenes, it is more suitable for object-centric images as it takes the underlying assumption that the scene has exactly the same semantics from any view, which is not true for large scenes with complex configurations due to view changes and occlusions. The text embedding learned from textual inversion is of the dimension of a single-world embedding, limiting its expressiveness in representing the subtleties complex contents.


%% file: figures/limitations.tex
\begin{figure}[t]
\centering
\begin{subfigure}[b]{\linewidth}
    \centering
    \includegraphics[width=.7\linewidth]{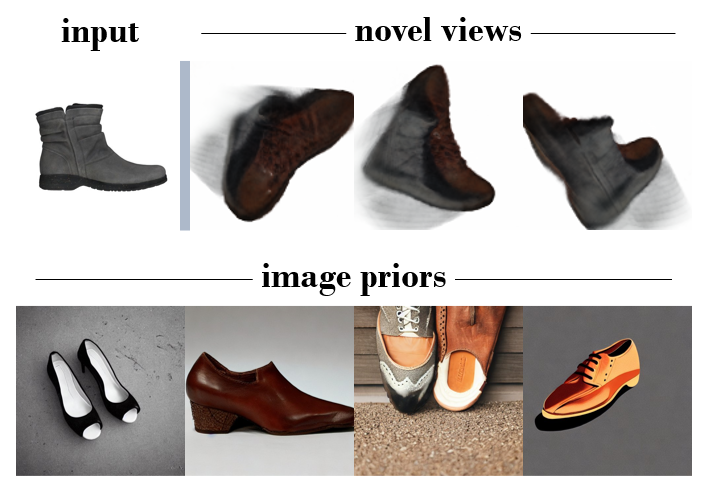}
    \caption{\textbf{A failure case due to the biases in the image diffusion model.}
    \textbf{Top:} Novel view synthesis results with text prompt \texttt{`a shoe in the style of <input>'.}
    \textbf{Bottom:} Images generated by \cite{Rombach_2022_CVPR} with text prompt ``a single shoe''. Yet half of the images have two shoes in it.
    }
    \label{fig:limitation_bias}
\end{subfigure}
\begin{subfigure}[b]{\linewidth}
    \centering
    \includegraphics[width=.7\linewidth]{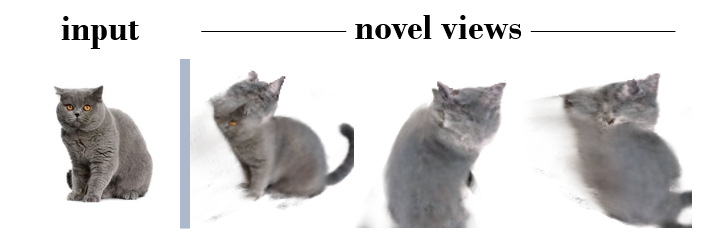}
    \caption{\textbf{A failure case on a highly deformable instance.}
    While the overall body shape of the cat is captured, the synthesized cat has two heads and two tails.
    }
    \label{fig:limitation_deformable}
\end{subfigure}
\caption{\textbf{Failure cases.}}
\vspace{-1.5em}
\label{fig:limitation}
\end{figure}

%% file: sections/X_appendix.tex
\appendix

\section{Additional Results}
\vspace{-.5em}

\input{figures_suppl/x_a_pumpkin}
\input{figures_suppl/more_results_wild}

Figure~\ref{suppl:fig:more_results_wild} shows our additional results and comparisons for images in the wild.
The results are presented in 4 groups, each group containing 3 objects from similar classes but with different content details and appearances.
We use this to test the capability of each method in capturing the overall semantics and visual feature variations from input images.

\paragraph{Comparison to DietNeRF~\cite{jain2021putting}}
\emph{For a fair comparison, DietNeRF is also optimized with the estimated depth map from the input image.}
While DietNeRF is able to maintain appearance consistency between different views, it fails to capture the overall geometry of the objects, especially when the object has complex geometric structures (such as the chairs in the 1st group, and the baskets in the 3rd group).
In the 4th group (the skirts), our generated textures form the unseen back regions are also closer to the input image than DietNeRF.

Our method also addresses the naturally existing ambiguity in novel-view inference, especially for the occluded regions in the input view.
For example, in the 3rd group in Figure~\ref{suppl:fig:more_results_wild}, the unseen spaces of the baskets are filled with different fruits/flowers/vegetables, instead of duplicating the input views as DietNeRF~\cite{jain2021putting}.
As a feature or as an inductive bias, such synthesis results are also affected by the 2D distribution from the image diffusion model.
For example, Figure \ref{suppl:fig:a_pumpkin} shows the image generation results by \cite{Rombach_2022_CVPR} with text prompt \texttt{`a pumpkin'}. Half of them are Jack-o'-lanterns. This makes our synthesized pumpkin also having the Jack-o'-lantern face at its back (the 3rd row of the 2nd group).

\paragraph{Comparison to SS3D \cite{vasudev2022ss3d}}
As a geometry-based method, SS3D captures better global geometries than DietNeRF even without the depth regularization, especially on the object classes covered by ShapeNet~\cite{chang2015shapenet} where the

%% file: figures_suppl/x_a_pumpkin.tex
\begin{figure}[t]
\centering
\includegraphics[width=.7\linewidth]{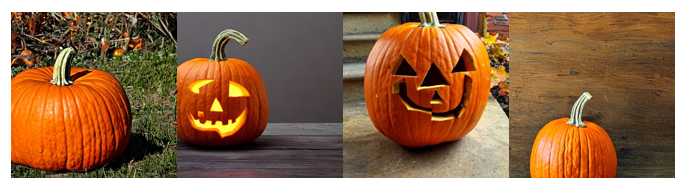}
\caption{Images generated by \cite{Rombach_2022_CVPR} with \texttt{`a pumpkin'}.
}
\vspace{-1em}
\label{suppl:fig:a_pumpkin}
\end{figure}

%% file: figures_suppl/more_results_wild.tex
\begin{figure*}[t]
\centering
\includegraphics[width=.98\linewidth]{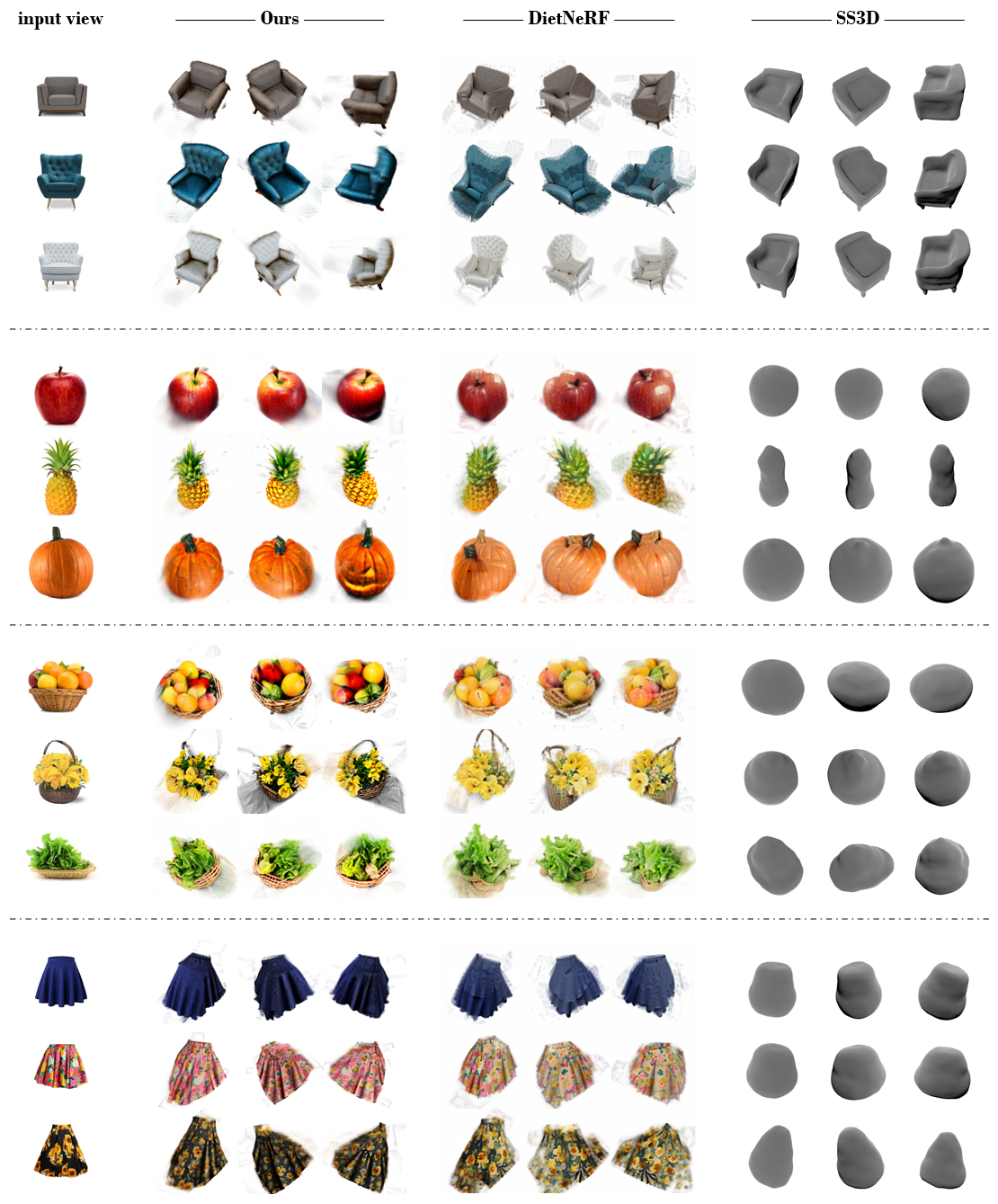}
\vspace{1cm}
\caption{\textbf{Additional results for images in the wild.}
}
\label{suppl:fig:more_results_wild}
\end{figure*}

%% file: PaperForReview.bbl
\begin{thebibliography}{10}\itemsep=-1pt

\bibitem{bautista2022gaudi}
Miguel~Angel Bautista, Pengsheng Guo, Samira Abnar, Walter Talbott, Alexander
  Toshev, Zhuoyuan Chen, Laurent Dinh, Shuangfei Zhai, Hanlin Goh, Daniel
  Ulbricht, et~al.
\newblock Gaudi: A neural architect for immersive 3d scene generation.
\newblock {\em arXiv preprint arXiv:2207.13751}, 2022.

\bibitem{chan2022efficient}
Eric~R Chan, Connor~Z Lin, Matthew~A Chan, Koki Nagano, Boxiao Pan, Shalini
  De~Mello, Orazio Gallo, Leonidas~J Guibas, Jonathan Tremblay, Sameh Khamis,
  et~al.
\newblock Efficient geometry-aware 3d generative adversarial networks.
\newblock In {\em Proceedings of the IEEE/CVF Conference on Computer Vision and
  Pattern Recognition}, pages 16123--16133, 2022.

\bibitem{chan2021pi}
Eric~R Chan, Marco Monteiro, Petr Kellnhofer, Jiajun Wu, and Gordon Wetzstein.
\newblock pi-gan: Periodic implicit generative adversarial networks for
  3d-aware image synthesis.
\newblock In {\em Proceedings of the IEEE/CVF conference on computer vision and
  pattern recognition}, pages 5799--5809, 2021.

\bibitem{chang2015shapenet}
Angel~X Chang, Thomas Funkhouser, Leonidas Guibas, Pat Hanrahan, Qixing Huang,
  Zimo Li, Silvio Savarese, Manolis Savva, Shuran Song, Hao Su, et~al.
\newblock Shapenet: An information-rich 3d model repository.
\newblock {\em arXiv preprint arXiv:1512.03012}, 2015.

\bibitem{chen2021mvsnerf}
Anpei Chen, Zexiang Xu, Fuqiang Zhao, Xiaoshuai Zhang, Fanbo Xiang, Jingyi Yu,
  and Hao Su.
\newblock Mvsnerf: Fast generalizable radiance field reconstruction from
  multi-view stereo.
\newblock In {\em Proceedings of the IEEE/CVF International Conference on
  Computer Vision}, pages 14124--14133, 2021.

\bibitem{chibane2021stereo}
Julian Chibane, Aayush Bansal, Verica Lazova, and Gerard Pons-Moll.
\newblock Stereo radiance fields (srf): Learning view synthesis for sparse
  views of novel scenes.
\newblock In {\em Proceedings of the IEEE/CVF Conference on Computer Vision and
  Pattern Recognition}, pages 7911--7920, 2021.

\bibitem{deng2022depth}
Kangle Deng, Andrew Liu, Jun-Yan Zhu, and Deva Ramanan.
\newblock Depth-supervised nerf: Fewer views and faster training for free.
\newblock In {\em Proceedings of the IEEE/CVF Conference on Computer Vision and
  Pattern Recognition}, pages 12882--12891, 2022.

\bibitem{downs2022google}
Laura Downs, Anthony Francis, Nate Koenig, Brandon Kinman, Ryan Hickman, Krista
  Reymann, Thomas~B McHugh, and Vincent Vanhoucke.
\newblock Google scanned objects: A high-quality dataset of 3d scanned
  household items.
\newblock {\em arXiv preprint arXiv:2204.11918}, 2022.

\bibitem{dupont2020equivariant}
Emilien Dupont, Miguel~Bautista Martin, Alex Colburn, Aditya Sankar, Josh
  Susskind, and Qi Shan.
\newblock Equivariant neural rendering.
\newblock In {\em International Conference on Machine Learning}, pages
  2761--2770. PMLR, 2020.

\bibitem{gal2022image}
Rinon Gal, Yuval Alaluf, Yuval Atzmon, Or Patashnik, Amit~H Bermano, Gal
  Chechik, and Daniel Cohen-Or.
\newblock An image is worth one word: Personalizing text-to-image generation
  using textual inversion.
\newblock {\em arXiv preprint arXiv:2208.01618}, 2022.

\bibitem{ho2020denoising}
Jonathan Ho, Ajay Jain, and Pieter Abbeel.
\newblock Denoising diffusion probabilistic models.
\newblock {\em Advances in Neural Information Processing Systems},
  33:6840--6851, 2020.

\bibitem{jain2021putting}
Ajay Jain, Matthew Tancik, and Pieter Abbeel.
\newblock Putting nerf on a diet: Semantically consistent few-shot view
  synthesis.
\newblock In {\em Proceedings of the IEEE/CVF International Conference on
  Computer Vision}, pages 5885--5894, 2021.

\bibitem{jensen2014large}
Rasmus Jensen, Anders Dahl, George Vogiatzis, Engin Tola, and Henrik Aan{\ae}s.
\newblock Large scale multi-view stereopsis evaluation.
\newblock In {\em Proceedings of the IEEE conference on computer vision and
  pattern recognition}, pages 406--413, 2014.

\bibitem{kim2022diffusionclip}
Gwanghyun Kim, Taesung Kwon, and Jong~Chul Ye.
\newblock Diffusionclip: Text-guided diffusion models for robust image
  manipulation.
\newblock In {\em Proceedings of the IEEE/CVF Conference on Computer Vision and
  Pattern Recognition}, pages 2426--2435, 2022.

\bibitem{lehar2003world}
Steven~M Lehar.
\newblock {\em The world in your head: A gestalt view of the mechanism of
  conscious experience}.
\newblock Psychology Press, 2003.

\bibitem{li2018differentiable}
Tzu-Mao Li, Miika Aittala, Fr{\'e}do Durand, and Jaakko Lehtinen.
\newblock Differentiable monte carlo ray tracing through edge sampling.
\newblock {\em ACM Transactions on Graphics (TOG)}, 37(6):1--11, 2018.

\bibitem{lin2014microsoft}
Tsung-Yi Lin, Michael Maire, Serge Belongie, James Hays, Pietro Perona, Deva
  Ramanan, Piotr Doll{\'a}r, and C~Lawrence Zitnick.
\newblock Microsoft coco: Common objects in context.
\newblock In {\em European conference on computer vision}, pages 740--755.
  Springer, 2014.

\bibitem{liu2022neural}
Yuan Liu, Sida Peng, Lingjie Liu, Qianqian Wang, Peng Wang, Christian Theobalt,
  Xiaowei Zhou, and Wenping Wang.
\newblock Neural rays for occlusion-aware image-based rendering.
\newblock In {\em Proceedings of the IEEE/CVF Conference on Computer Vision and
  Pattern Recognition}, pages 7824--7833, 2022.

\bibitem{luo2021diffusion}
Shitong Luo and Wei Hu.
\newblock Diffusion probabilistic models for 3d point cloud generation.
\newblock In {\em Proceedings of the IEEE/CVF Conference on Computer Vision and
  Pattern Recognition}, pages 2837--2845, 2021.

\bibitem{meng2021sdedit}
Chenlin Meng, Yang Song, Jiaming Song, Jiajun Wu, Jun-Yan Zhu, and Stefano
  Ermon.
\newblock Sdedit: Image synthesis and editing with stochastic differential
  equations.
\newblock {\em arXiv preprint arXiv:2108.01073}, 2021.

\bibitem{meng2021gnerf}
Quan Meng, Anpei Chen, Haimin Luo, Minye Wu, Hao Su, Lan Xu, Xuming He, and
  Jingyi Yu.
\newblock Gnerf: Gan-based neural radiance field without posed camera.
\newblock In {\em Proceedings of the IEEE/CVF International Conference on
  Computer Vision}, pages 6351--6361, 2021.

\bibitem{mi2022im2nerf}
Lu Mi, Abhijit Kundu, David Ross, Frank Dellaert, Noah Snavely, and Alireza
  Fathi.
\newblock im2nerf: Image to neural radiance field in the wild.
\newblock {\em arXiv preprint arXiv:2209.04061}, 2022.

\bibitem{mildenhall2021nerf}
Ben Mildenhall, Pratul~P Srinivasan, Matthew Tancik, Jonathan~T Barron, Ravi
  Ramamoorthi, and Ren Ng.
\newblock Nerf: Representing scenes as neural radiance fields for view
  synthesis.
\newblock {\em Communications of the ACM}, 65(1):99--106, 2021.

\bibitem{mitra2009shadow}
Niloy~J Mitra and Mark Pauly.
\newblock Shadow art.
\newblock {\em ACM Transactions on Graphics}, 28(CONF):156--1, 2009.

\bibitem{muller2022instant}
Thomas M{\"u}ller, Alex Evans, Christoph Schied, and Alexander Keller.
\newblock Instant neural graphics primitives with a multiresolution hash
  encoding.
\newblock {\em arXiv preprint arXiv:2201.05989}, 2022.

\bibitem{niemeyer2022regnerf}
Michael Niemeyer, Jonathan~T Barron, Ben Mildenhall, Mehdi~SM Sajjadi, Andreas
  Geiger, and Noha Radwan.
\newblock Regnerf: Regularizing neural radiance fields for view synthesis from
  sparse inputs.
\newblock In {\em Proceedings of the IEEE/CVF Conference on Computer Vision and
  Pattern Recognition}, pages 5480--5490, 2022.

\bibitem{niemeyer2021giraffe}
Michael Niemeyer and Andreas Geiger.
\newblock Giraffe: Representing scenes as compositional generative neural
  feature fields.
\newblock In {\em Proceedings of the IEEE/CVF Conference on Computer Vision and
  Pattern Recognition}, pages 11453--11464, 2021.

\bibitem{poole2022dreamfusion}
Ben Poole, Ajay Jain, Jonathan~T Barron, and Ben Mildenhall.
\newblock Dreamfusion: Text-to-3d using 2d diffusion.
\newblock {\em arXiv preprint arXiv:2209.14988}, 2022.

\bibitem{qin2022}
Xuebin Qin, Hang Dai, Xiaobin Hu, Deng-Ping Fan, Ling Shao, and Luc~Van Gool.
\newblock Highly accurate dichotomous image segmentation.
\newblock In {\em ECCV}, 2022.

\bibitem{radford2021learning}
Alec Radford, Jong~Wook Kim, Chris Hallacy, Aditya Ramesh, Gabriel Goh,
  Sandhini Agarwal, Girish Sastry, Amanda Askell, Pamela Mishkin, Jack Clark,
  et~al.
\newblock Learning transferable visual models from natural language
  supervision.
\newblock In {\em International Conference on Machine Learning}, pages
  8748--8763. PMLR, 2021.

\bibitem{radford2019language}
Alec Radford, Jeffrey Wu, Rewon Child, David Luan, Dario Amodei, Ilya
  Sutskever, et~al.
\newblock Language models are unsupervised multitask learners.
\newblock {\em OpenAI blog}, 1(8):9, 2019.

\bibitem{ramesh2022hierarchical}
Aditya Ramesh, Prafulla Dhariwal, Alex Nichol, Casey Chu, and Mark Chen.
\newblock Hierarchical text-conditional image generation with clip latents.
\newblock {\em arXiv preprint arXiv:2204.06125}, 2022.

\bibitem{ranftl2021vision}
Ren{\'e} Ranftl, Alexey Bochkovskiy, and Vladlen Koltun.
\newblock Vision transformers for dense prediction.
\newblock In {\em Proceedings of the IEEE/CVF International Conference on
  Computer Vision}, pages 12179--12188, 2021.

\bibitem{roessle2022dense}
Barbara Roessle, Jonathan~T Barron, Ben Mildenhall, Pratul~P Srinivasan, and
  Matthias Nie{\ss}ner.
\newblock Dense depth priors for neural radiance fields from sparse input
  views.
\newblock In {\em Proceedings of the IEEE/CVF Conference on Computer Vision and
  Pattern Recognition}, pages 12892--12901, 2022.

\bibitem{Rombach_2022_CVPR}
Robin Rombach, Andreas Blattmann, Dominik Lorenz, Patrick Esser, and Bj\"orn
  Ommer.
\newblock High-resolution image synthesis with latent diffusion models.
\newblock In {\em Proceedings of the IEEE/CVF Conference on Computer Vision and
  Pattern Recognition (CVPR)}, pages 10684--10695, June 2022.

\bibitem{saharia2022palette}
Chitwan Saharia, William Chan, Huiwen Chang, Chris Lee, Jonathan Ho, Tim
  Salimans, David Fleet, and Mohammad Norouzi.
\newblock Palette: Image-to-image diffusion models.
\newblock In {\em ACM SIGGRAPH 2022 Conference Proceedings}, pages 1--10, 2022.

\bibitem{saharia2022photorealistic}
Chitwan Saharia, William Chan, Saurabh Saxena, Lala Li, Jay Whang, Emily
  Denton, Seyed Kamyar~Seyed Ghasemipour, Burcu~Karagol Ayan, S~Sara Mahdavi,
  Rapha~Gontijo Lopes, et~al.
\newblock Photorealistic text-to-image diffusion models with deep language
  understanding.
\newblock {\em arXiv preprint arXiv:2205.11487}, 2022.

\bibitem{schuhmann2021laion}
Christoph Schuhmann, Richard Vencu, Romain Beaumont, Robert Kaczmarczyk,
  Clayton Mullis, Aarush Katta, Theo Coombes, Jenia Jitsev, and Aran
  Komatsuzaki.
\newblock Laion-400m: Open dataset of clip-filtered 400 million image-text
  pairs.
\newblock {\em arXiv preprint arXiv:2111.02114}, 2021.

\bibitem{schwarz2020graf}
Katja Schwarz, Yiyi Liao, Michael Niemeyer, and Andreas Geiger.
\newblock Graf: Generative radiance fields for 3d-aware image synthesis.
\newblock {\em Advances in Neural Information Processing Systems},
  33:20154--20166, 2020.

\bibitem{simonyan2014very}
Karen Simonyan and Andrew Zisserman.
\newblock Very deep convolutional networks for large-scale image recognition.
\newblock {\em arXiv preprint arXiv:1409.1556}, 2014.

\bibitem{song2020denoising}
Jiaming Song, Chenlin Meng, and Stefano Ermon.
\newblock Denoising diffusion implicit models.
\newblock {\em arXiv preprint arXiv:2010.02502}, 2020.

\bibitem{song2019generative}
Yang Song and Stefano Ermon.
\newblock Generative modeling by estimating gradients of the data distribution.
\newblock {\em Advances in Neural Information Processing Systems}, 32, 2019.

\bibitem{song2020score}
Yang Song, Jascha Sohl-Dickstein, Diederik~P Kingma, Abhishek Kumar, Stefano
  Ermon, and Ben Poole.
\newblock Score-based generative modeling through stochastic differential
  equations.
\newblock {\em arXiv preprint arXiv:2011.13456}, 2020.

\bibitem{trevithick2021grf}
Alex Trevithick and Bo Yang.
\newblock Grf: Learning a general radiance field for 3d representation and
  rendering.
\newblock In {\em Proceedings of the IEEE/CVF International Conference on
  Computer Vision}, pages 15182--15192, 2021.

\bibitem{vasudev2022ss3d}
Kalyan~Alwala Vasudev, Abhinav Gupta, and Shubham Tulsiani.
\newblock Pre-train, self-train, distill: A simple recipe for supersizing 3d
  reconstruction.
\newblock In {\em Computer Vision and Pattern Recognition (CVPR)}, 2022.

\bibitem{wang2021ibrnet}
Qianqian Wang, Zhicheng Wang, Kyle Genova, Pratul~P Srinivasan, Howard Zhou,
  Jonathan~T Barron, Ricardo Martin-Brualla, Noah Snavely, and Thomas
  Funkhouser.
\newblock Ibrnet: Learning multi-view image-based rendering.
\newblock In {\em Proceedings of the IEEE/CVF Conference on Computer Vision and
  Pattern Recognition}, pages 4690--4699, 2021.

\bibitem{wang2004image}
Zhou Wang, Alan~C Bovik, Hamid~R Sheikh, and Eero~P Simoncelli.
\newblock Image quality assessment: from error visibility to structural
  similarity.
\newblock {\em IEEE transactions on image processing}, 13(4):600--612, 2004.

\bibitem{wang2021nerf}
Zirui Wang, Shangzhe Wu, Weidi Xie, Min Chen, and Victor~Adrian Prisacariu.
\newblock Nerf--: Neural radiance fields without known camera parameters.
\newblock {\em arXiv preprint arXiv:2102.07064}, 2021.

\bibitem{watson2022novel}
Daniel Watson, William Chan, Ricardo Martin-Brualla, Jonathan Ho, Andrea
  Tagliasacchi, and Mohammad Norouzi.
\newblock Novel view synthesis with diffusion models.
\newblock {\em arXiv preprint arXiv:2210.04628}, 2022.

\bibitem{yariv2020multiview}
Lior Yariv, Yoni Kasten, Dror Moran, Meirav Galun, Matan Atzmon, Basri Ronen,
  and Yaron Lipman.
\newblock Multiview neural surface reconstruction by disentangling geometry and
  appearance.
\newblock {\em Advances in Neural Information Processing Systems},
  33:2492--2502, 2020.

\bibitem{ye2021shelf}
Yufei Ye, Shubham Tulsiani, and Abhinav Gupta.
\newblock Shelf-supervised mesh prediction in the wild.
\newblock In {\em Proceedings of the IEEE/CVF Conference on Computer Vision and
  Pattern Recognition}, pages 8843--8852, 2021.

\bibitem{yu2021pixelnerf}
Alex Yu, Vickie Ye, Matthew Tancik, and Angjoo Kanazawa.
\newblock pixelnerf: Neural radiance fields from one or few images.
\newblock In {\em Proceedings of the IEEE/CVF Conference on Computer Vision and
  Pattern Recognition}, pages 4578--4587, 2021.

\bibitem{zhang2020path}
Cheng Zhang, Bailey Miller, Kan Yan, Ioannis Gkioulekas, and Shuang Zhao.
\newblock Path-space differentiable rendering.
\newblock {\em ACM transactions on graphics}, 39(4), 2020.

\bibitem{zhang2019differential}
Cheng Zhang, Lifan Wu, Changxi Zheng, Ioannis Gkioulekas, Ravi Ramamoorthi, and
  Shuang Zhao.
\newblock A differential theory of radiative transfer.
\newblock {\em ACM Transactions on Graphics (TOG)}, 38(6):1--16, 2019.

\bibitem{zhang2021path}
Cheng Zhang, Zihan Yu, and Shuang Zhao.
\newblock Path-space differentiable rendering of participating media.
\newblock {\em ACM Transactions on Graphics (TOG)}, 40(4):1--15, 2021.

\bibitem{zhang2018unreasonable}
Richard Zhang, Phillip Isola, Alexei~A Efros, Eli Shechtman, and Oliver Wang.
\newblock The unreasonable effectiveness of deep features as a perceptual
  metric.
\newblock In {\em Proceedings of the IEEE conference on computer vision and
  pattern recognition}, pages 586--595, 2018.

\bibitem{zhou20213d}
Linqi Zhou, Yilun Du, and Jiajun Wu.
\newblock 3d shape generation and completion through point-voxel diffusion.
\newblock In {\em Proceedings of the IEEE/CVF International Conference on
  Computer Vision}, pages 5826--5835, 2021.

\end{thebibliography}
